\documentclass[10pt,twocolumn,letterpaper]{article}

\usepackage{iccv}
\usepackage{times}
\usepackage{epsfig}
\usepackage{graphicx}
\usepackage{amsmath}
\usepackage{amssymb}
\usepackage[breaklinks=true,bookmarks=false]{hyperref}

\iccvfinalcopy % *** Uncomment this line for the final submission

\def\iccvPaperID{****} % *** Enter the ICCV Paper ID here

\begin{document}

%%%%%%%%% TITLE
\title{End-to-End Learning Deep CRF models for Multi-Object Tracking}

\author{Jun Xiang$^1$  \enspace  Chao Ma$^2$  \enspace Guohan Xu$^1$ \enspace Jianhua Hou$^1$ \\
	\small 	$^1$South-Central University for Nationalities, Wuhan, China \qquad
	\small 	$^2$Shanghai Jiao Tong University, Shanghai,China\\
	{\tt\small \{junxiang, zil\}@scuec.edu.cn}, {\tt\small chaoma99@gmai.com}
}

%\author{
%        Jun Xiang\\
%		South-Central University for Nationalities\\
%		Wuhan, China\\
%		{\tt\small junxiang@scuec.edu.cn}
%		% For a paper whose authors are all at the same institution,
%		% omit the following lines up until the closing ``}''.
%		% Additional authors and addresses can be added with ``\and'',
%		% just like the second author.
%		% To save space, use either the email address or home page, not both
%		\and
%		Chao Ma\\
%		Shanghai Jiao Tong University\\
%		Shanghai,China\\
%		{\tt\small chaoma99@gmai.com}
%        \and
%		Guohan Xu,Jinahua Hou\\
%		South-Central University for Nationalities\\
%		Wuhan,China\\
%	   %\and
%		%Jianhua Hou\\
%%    South-Central University for Nationalities\\
%%		Wuhan,China\\
%%        {\tt\small zil@scuec.edu.cn}
%	}

\maketitle

    \begin{abstract}
		Existing deep multi-object tracking (MOT) approaches first learn a deep representation to describe target objects and then associate detection results by optimizing a linear assignment problem. Despite demonstrated successes, it is challenging to discriminate target objects under mutual occlusion or to reduce identity switches in crowded scenes.  In this paper, we propose learning deep conditional random field (CRF) networks, aiming to model the assignment costs as unary potentials and the long-term dependencies among detection results as pairwise potentials. Specifically, we use a bidirectional long short-term memory (LSTM) network to encode the long-term dependencies. We pose the CRF inference as a recurrent neural network learning process using the standard gradient descent algorithm, where unary and pairwise potentials are jointly optimized in an end-to-end manner. Extensive experimental results on the challenging MOT datasets including MOT-2015 and MOT-2016, demonstrate that our approach achieves the state of the art performances in comparison with published works on both benchmarks.
	\end{abstract}
	
	%%%%%%%%% BODY TEXT
	\section{Introduction}
	
	Multi-object tracking (MOT) aims at finding the trajectories of multiple target objects simultaneously while maintaining their identities. The performance of MOT approaches hinges on an effective representation of target objects as well as robust data association to assign detection results to corresponding trajectories.
	
\begin{figure}[t]
		\begin{center}
			\includegraphics[width=84mm]{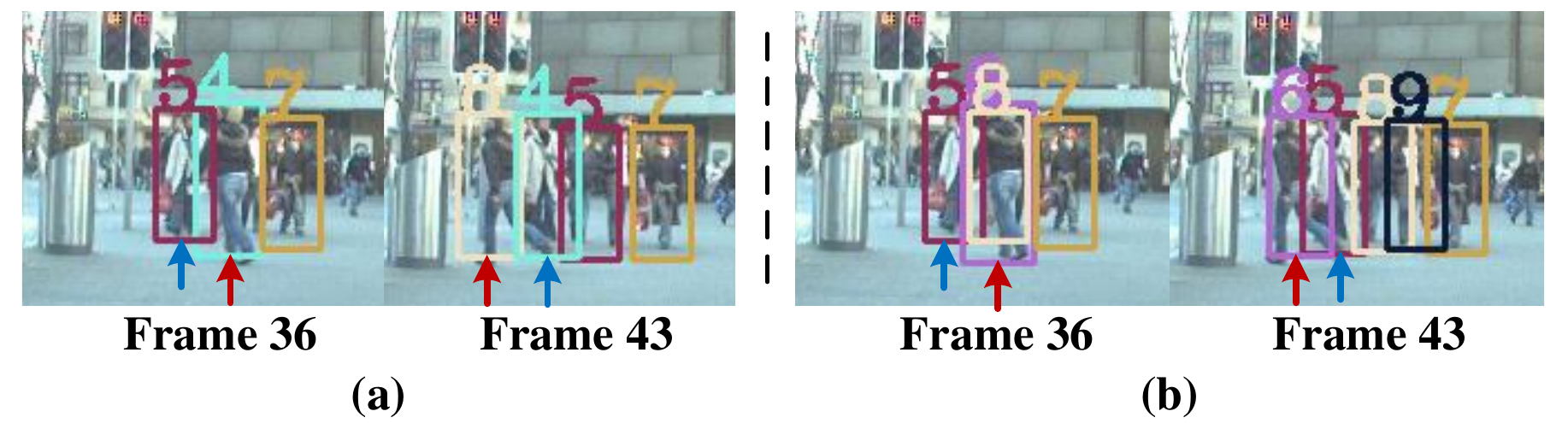}
		\end{center}
		\vspace{-5mm}
		\caption{Illustration of the importance of modeling the dependencies between target objects. (a) The state-of-the-art deep MOT method~\cite{siamesecnn2016LearningBT}  does not exploit the dependencies between targets and fails to maintain a consistent identity for target 4 and target 5. (b) Our method built on deep CRF models succeeds in tracking the target 5 and target 6 over time.}
		\label{fig:Fig1}
		\vspace{-5mm}
	\end{figure}

	Existing deep MOT methods put more emphasis on learning a favorable representation to describe target objects and associate detection results by reasoning a linear assignment cost~\cite{NHTKim2015MultipleHT,QuadrupletSon2017MultiobjectTW,Kuan2017Recurrent,siamesecnn2016LearningBT,ConstrainedWang2016JointLO}. Despite noticeable progress, existing deep MOT methods are unlikely to distinguish target objects in crowded scenes, where occlusions, noisy detections and large appearance variations often occur, due to the lack of the component for modeling long-term dependencies among detection results over time. In MOT, the dependencies among targets mean the structural correlation of targets in the spatiotemporal domain, which are subject to a number of physical constraints. For instance, a target object cannot be at two places at one time, and two target objects cannot occupy the same space at the same time. %Sometimes, targets may choose to move together as a group for a while or pushes apart to avoid a collision.  And
	Taking these constraints into consideration in data association helps to improve MOT performance, and  some visual results shown in Fig.~\ref{fig:Fig1}. Without considering the dependencies among multiple targets, the state-of-the-art deep MOT method~\cite{siamesecnn2016LearningBT} fails to maintain the same identities such as targets marked with color arrows,  whereas our method succeeds in tracking them by considering these constraints.	

	%(a) shows a tracking failure example by Siamese CNN to association detections independently, when a little girl in write makes a sharp direction change which leads to failure of the appearance model. In Fig.\ref{fig:Fig1}(b), by incorporating pairwise terms, we are able to locate the nonlinear path and to track the girl correctly by following the state consistency constraints(see section X for details).
	
	In the literature, MOT approaches~\cite{Bo2014Multi,Near-OnlineChoi2015NearOnlineMT,Heili2014Exploiting,HierarchicalHuang2008RobustOT,HFRFXiang2016MultitargetTU,Discrete-Continuous2016MultiTargetTB,networkflowsLin2008GlobalDA} before the prevalence of deep learning widely use conditional random field (CRF) to model the dependencies among targets as the pairwise potentials explicitly~\cite{Bo2014Multi,Heili2014Exploiting,Learningaffinities2011LearningAA}.
	%{\color{red}Conditional random field (CRF) is well-known for its ability to exploit data structure and has been widely used for modeling long-term dependence among the detections/tracklets in pre-deep learning MOT literature~\cite{Bo2014Multi,Heili2014Exploiting,Learningaffinities2011LearningAA,Heili2011Detection,Near-OnlineChoi2015NearOnlineMT}.}
	%
	However, MOT typically involves varying number of targets, %the constraint that two targets cannot occupy the same space simultaneously,
	which makes CRF inference intractable. Existing approaches built on CRF have to resort to heuristic ~\cite{Bo2014Multi,Learningaffinities2011LearningAA} or iterative approximate algorithms ~\cite{Heili2014Exploiting,Heili2011Detection}.
%	{\color{red}However, due to the complex structure of the CRF models for multitarget tracking, the parameter learning and inference of such models are often intractable, and hence conducted using a heuristic algorithm  ~\cite{Bo2014Multi,Learningaffinities2011LearningAA}  or an iterative approximate algorithm ~\cite{Heili2014Exploiting,Heili2011Detection}.}
	%
	Moreover, these approaches assume that multiple targets are subject to Gaussian distribution~\cite{Bo2014Multi,Heili2014Exploiting} when formulating the CRF potentials, meaning that multiple targets share the same states. This is in contrast to the fact that there are both consistent and repellent dependencies between moving targets.

	In this work, we propose end-to-end learning CRF for multi-object tracking within a unified deep architecture. The proposed method mainly consists of two steps: (i) designing task-specific networks to implicitly learn unary and pairwise potentials respectively; (ii) conducting CRF inference as a recurrent neural network learning process using the standard gradient descent algorithm.
	Our work partially resembles the recent CRF-RNN framework for semantic segmentation~\cite{CRFasRNNZheng2015ConditionalRF,ArbitraryLarsson2017LearningAP}, where the inference of CRF model is embedded into a recurrent neural network as well. However, %directly applying the CRF-RNN model to MOT does not work as
	the noticeable difference lies in that CRF-RNN is learned for determined pixel labeling rather than for graphical models with varying nodes (\ie, the number of targets are varying from frame to frame in MOT). We model the unary terms by a matching network to compute the assignment cost. To fully exploit dependencies between detection results (\ie, CRF nodes), we pay more attention to difficult node pairs and formulate the labeling of these pairs as a joint probability matching problem. To be specific, we use a bidirectional LSTM to implicitly learn the pairwise terms. Finally, we pose the overall CRF inference as a recurrent neural network learning process using the standard gradient descent algorithm, where unary and pairwise potentials are jointly optimized in an end-to-end manner.

 %it is non-trivial to apply the same strategy to MOT problem, since the CRF-RNN learned for pixel-level labeling task cannot be directly employed in graphical models with different structure (e.g. unknown and varying number of targets in MOT). In this work, the unary terms are learned by CNN to model the assignment costs for global affinity. To fully exploit dependencies between detections (between CRF nodes) , we define difficult node pairs and formulate labeling of these pairs as a joint (probability) matching problem. In this way, pairwise terms could be learned implicitly by a highly expressive bidirectional LSTM. Finally, We pose the CRF inference as a recurrent neural network learning process using the standard gradient descent algorithm, where unary and pairwise potentials are jointly optimized in an end-to-end manner.
	
	In summary, the main contributions of this work are as follows:
	\begin{itemize}
		\itemsep 0mm
		\item We make the first attempt to learn deep CRF models for multi-object tracking in an end-to-end manner.
		\item We propose a bidirectional LSTM to learn pairwise potentials that encode the long-term dependencies between CRF nodes.
		\item We extensively evaluate our method on the MOT-2015, MOT-2016 datasets. Our method achieves new state-of-the-art performance on both datasets.
	\end{itemize}

	\section{Related Work}
	In the tracking-by-detection framework, multi-object tracking consists of two critical components: a matching metric to compute assignment costs and data association. We briefly review the related work with the use of CRF and deep learning.
	
	\vspace{-3mm}
	\paragraph{MOT with CRF.}  As a classical graphical model, CRFs have been used extensively for solving the assignment problem in multi-object tracking~\cite{Bo2014Multi,Heili2014Exploiting,Learningaffinities2011LearningAA,Heili2011Detection,Near-OnlineChoi2015NearOnlineMT}. CRFs relax the independent assumption among detections, aiming at using the long-term dependence among targets to better distinguish targets in crowded scenes. Yang et al.~\cite{Learningaffinities2011LearningAA}adopt a CRF model and consider both tracklet affinities and dependencies among them, which are represented by unary term costs and pairwise term costs respectively. The approach in~\cite{Bo2014Multi} further considers distinguishing difficult pairs of targets. While unary potentials are based on motion and appearance models for discriminating all targets, pairwise ones are designed for differentiating spatially close tracklets. Heili et al.~\cite{Heili2014Exploiting,Heili2011Detection} perform association at the detection level. To exploit long-term connectivity between pairs of detections, the CRF is formulated in terms of similarity/dissimilarity pairwise factors and additional higher-order potentials defined in terms of label costs. In~\cite{Discrete-Continuous2016MultiTargetTB} , the energy term of CRF is augmented with a continuous component to jointly solve the discrete data association and continuous trajectory estimation problems. %and the dependency model includes an observation likelihood and several physically
	%motivated priors, such as the target's dynamics, exclusion
	%and trajectory persistence. {\color{red}The main drawback of existing methods is that they are limited to  hand-designed force terms.}
	%
	%
	Note that in MOT, the CRF inference is intractable as proved by Heili~\cite{Heili2014Exploiting} that an energy function in MOT that only containing unary and pairwise terms does not follow the submodularity or regularity principle, and hence cannot be solved using standard optimization techniques like graph cuts.
	Unlike  existing approaches using heuristic~\cite{Bo2014Multi} or iterative approximate algorithms~\cite{Heili2014Exploiting,Heili2011Detection}, we propose to learn deep CRF models as recurrent neural networks in an end-to-end manner.

	\vspace{-3mm}
	\paragraph{MOT with Deep Learning.} Recently, deep learning has been applied to multi-object tracking~\cite{OnlineRNNMilan2017OnlineMT,siamesecnn2016LearningBT,Sadeghian2017TrackingTU,ConstrainedWang2016JointLO,NHTKim2015MultipleHT,QuadrupletSon2017MultiobjectTW,PrincipledIntegrationBeyer2017TowardsAP,LiftedMulticutTang2017MultiplePT}. The trend in this line is to learn deep representations~\cite{networkflowsLin2008GlobalDA,Globally-optimalgreedy2011GloballyoptimalGA,CoherentDynamicsWang2017TrackletAB,QuadrupletSon2017MultiobjectTW}, and then employ traditional assignment strategies such as bipartite matching~\cite{QuadrupletSon2017MultiobjectTW},  or linear assignment for optimization~\cite{ConstrainedWang2016JointLO}. Leal-Taixet al.~\cite{siamesecnn2016LearningBT}  propose a Siamese CNN to estimate the similarity between targets and detections. Tang et al.~\cite{LiftedMulticutTang2017MultiplePT} go a step further by treating MOT as the person Re-ID problem and develop a Siamese ID-Net to compute association costs between detections. %, where networks for re-identifying persons are designed and trained by fusing human visual and pose information.
	Sadeghian et al.~\cite{Sadeghian2017TrackingTU} exploit CNN and LSTM to build the affinity measures based on appearance, motion and interaction cues. Deep metric learning is also proposed for representation learning, so that the affinity model is robust to appearance variations~\cite{QuadrupletSon2017MultiobjectTW,bmvcXiang2018MultipleOT}.
	While considerable attention has been paid to DNN-based assignment problems, the optimization with respect to the dependencies between targtes is less investigated. In this work, we pose the assignment optimization as a recurrent neural network learning process, where unary and pairwise potentials are jointly optimized in an end-to-end manner. %Furthermore, our pairwise terms are based on difficult node pairs that encode long-term dependencies among tracklets pairs.
	Our work shares a similar motivation with CRF-RNN approaches~\cite{CRFasRNNZheng2015ConditionalRF,ArbitraryLarsson2017LearningAP} for semantic segmentation. The significant difference lies in that CRF-RNN is learned for determined pixel labeling (\ie, the number of labels and pixels are given), whereas our method deals with dynamic graphical models with varying number of nodes (targets) for MOT.
	
	\begin{figure*}
		\begin{center}
			\includegraphics[width=170mm]{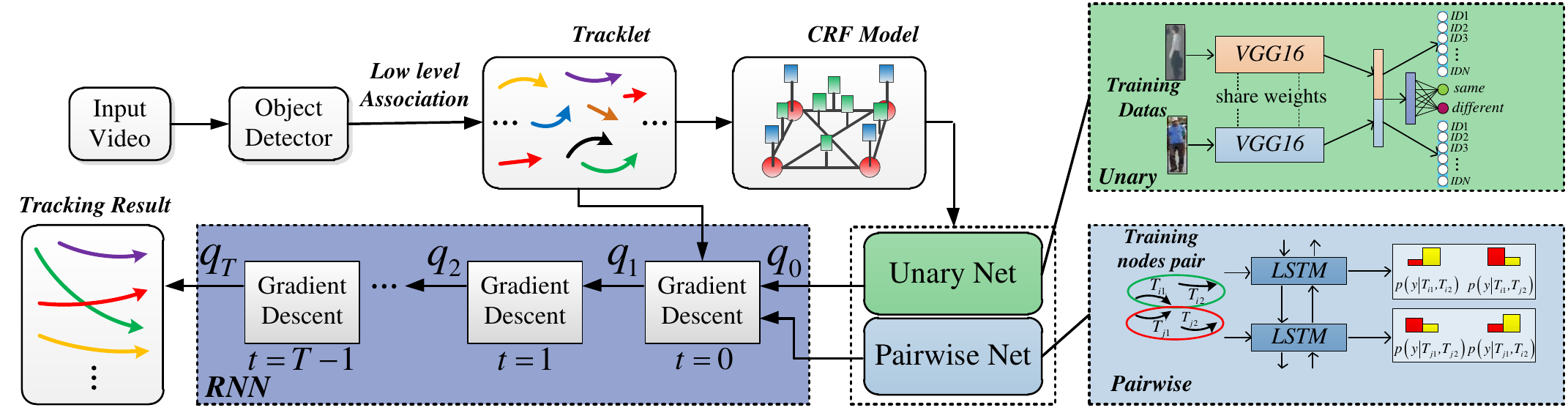}
		\end{center}
        \vspace{-3mm}
		\caption{Method overview. We employ a two-threshold strategy to connect detection results into short but reliable tracklets. We formulate tracklet association as a labeling problem in the Conditional Random Field (CRF) framework and perform CRF inference by learning a deep recurrent neural network end-to-end.}
		\label{fig:Fig2}
    \vspace{-2mm}
	\end{figure*}

%%%%%%%%%%%%%%%%%%%%%%%%%%%%%%%%%%%%%%%%%%%%%%%%%xjxjxjxjxj%%%%%%%%%%%%%%%%%%%%%%%%%%%%%%%%%%%%%%%
    \section{Proposed Method}
 In this section, we first describe how the data association based MOT is mapped into a CRF labeling problem. The designs of unary and pairwise potentials are then introduced. We present RNN based inference strategy and finally the overview of entire structure.
    \subsection{Problem Formulation}
Similar to~\cite{onlineCRFYang2012AnOL}, we treat the task of MOT as a CRF labeling problem. The flowchart of our approach is shown in Fig.~\ref{fig:Fig2}. We use noisy detections as a starting point, then a two-threshold strategy ~\cite{HierarchicalHuang2008RobustOT} is employed to connect detection responses into short but reliable tracklets. A tracklet ${T_i} = \{ d_i^{t_i^s}, \cdots, d_i^{t_i^e}\}$ represents a set of detection responses in consecutive frames, starting at frame $t_i^s$ and ending at $t_i^e$. Let $d_i^t = \{ \;p_i^t,\;s_i^t\;\}$ denotes a detection of target $T_i$ at time step $t$, with position $p_i^t$ and size $s_i^t$.

A graph ${\cal G} = (\;{\cal V},\,{\cal E}\,)$ for a CRF is created over the set of tracklets ${\cal T}{\rm{ = \{ }}{T_i}{\rm{\} }}$, where ${\cal V}$ and ${\cal E}$ denote the set of nodes and edges, respectively. Tracklet $T_i$ is linkable to $T_j$ if the gap between the end of $T_i$ and the beginning of $T_j$ satisfies $0 < t_j^s - t_i^e < {T_{thr}}$, where $T_{thr}$ is a threshold for maximum gap between any linkable tracklet pair. A linkable pair of tracklets $({T_i}_1 \to {T_i}_2)$ forms the graph node ${v_i} = ({T_i}_1 \to {T_i}_2),\;\;i = 1,2, \cdots |{\cal V}|$. An edge ${e_j} = ({v_{j1}}\;,{v_{j2}}) \in {\cal E}$ represents a correlation between a pair of two nodes. Furthermore, each node is associated with a binary label variable ${x_i} \in {\cal L} = \{ 0,\,\,1\}$, where ${x_i} = 1$ indicates the two tracklets in node $v_i$ should be associated and $x_i=0$ means the opposite.

Then $X = [{x_1},{x_{2\;}}, \cdots {x_{|{\cal V}|\;}}]$, a variable realization over all the nodes, corresponds to an association hypothesis of the set of tracklets ${\cal T}$. The pair of $(X,{\cal T})$ is modeled as a CRF characterized by the Gibbs distribution, formulated as
\begin{equation}	
P(X = \mathbf{\mathbf{x}}|{\cal T}) = \frac{1}{{Z({\cal T})}}\exp ( - E(\mathbf{x}|{\cal T}))
\label{Eq:eq1}
\end{equation}
where $E(\mathbf{x}|{\cal T})$ denotes the Gibbs energy function with respect to the labeling $\mathbf{x} \in {{\cal L}^{|{\cal V}|}}$, $Z({\cal T})$ is the partition function. For simplicity of notation, the conditioning on ${\cal T}$ will from now on be dropped. And the normalization term $Z({\cal T})$ in Eq.~\ref{Eq:eq1} can be omitted when solving for the maximum-probability labeling $\mathbf{x}$ for a particular set of observations ${\cal T}$. In this paper, energy function is restricted containing a set of unary and pairwise terms and formulated as:
\begin{equation}	
E(\mathbf{x}) = \sum\limits_{i \in {\cal V}} {{\varphi _i}({x_i})}  + \sum\limits_{(i,j) \in {\cal E}} {{\varphi _{ij}}({x_i},{x_j})}
\label{Eq:eq2}
\end{equation}
where ${\varphi _i}:{\cal L} \to \mathbb{R}$ and ${\varphi _{ij}}:{\cal L} \times {\cal L} \to \mathbb{R}$ are the unary and pairwise potentials, respectively. As is common with CRFs, our tracking problem is transformed into an energy minimization problem.
    \subsection{Unary Potentials}
The unary potential ${\varphi _i}({x_i})$ specifies the energy cost of assigning label $x_i$ to node $v_i$. In this paper we obtain our unary from a Siamese CNN with shared parameters. Roughly speaking, the Siamese Net as a appearance descriptor outputs a probability of a linkable tracklet pair containing each label class (i.e. belonging to the same target or not). Denoting the probability for node $v_i$ and label $x_i$ as ${z_{i:{x_i}}}$, the unary potential is
\begin{equation}
{\varphi _i}({x_i}) =  - {\omega _u}\log ({z_{i:{x_i}}} + \varepsilon )
\label{Eq:eq3}
\end{equation}
where ${\omega _u}$ is a parameter controlling the impact of the unary potentials and $\varepsilon$ is introduced to avoid numerical problems for small values of ${z_{i:{x_i}}}$.

    \subsection{Pairwise potential}
The pairwise potential ${\varphi _{ij}}({x_i},{x_j})$ is defined as the energy cost of assigning label $x_i$ to node $v_i$ and $x_j$ to $v_j$ simultaneously. To address the challenge of association in crowded scenes, we propose a novel potential function built on difficult node pairs, and limit the number of edges by imposing pairwise terms on these pairs for efficiency.
\begin{figure}
\begin{center}
\includegraphics[width=84mm]{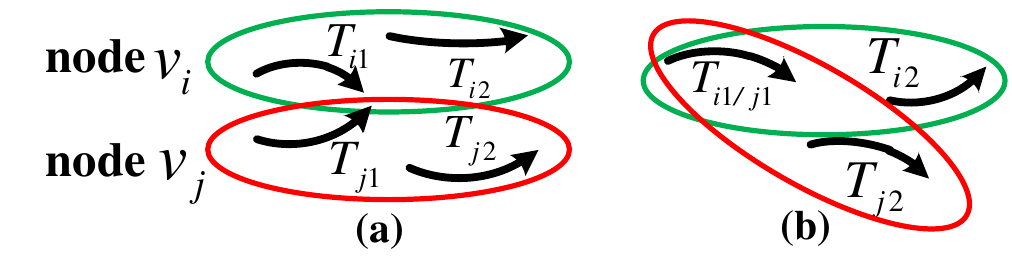}
\end{center}
\vspace{-5mm}
\caption{Two examples of difficult node pairs. (a) The tracklet $T_{i1}$ is tail-close to $T_{j1}$. (b) The node pair  shares the same tracklet.}
\label{fig:Fig3}
\vspace{-5mm}
\end{figure}

\vspace{-3mm}
\paragraph{Difficult Node Pairs.} Difficult node pairs are defined as spatio-temporal head-close or tail-close tracklet pairs like in~\cite{onlineCRFYang2012AnOL}, which are usually caused by close tracklets crossing each other or by two targets interacting as a group. In this situation, we impose consistency constraints to restrict the labeling of difficult node pairs.

Take Fig.~\ref{fig:Fig3}(a) as a example. Since node ${v_i} = ({T_{i1}} \to {T_{i2}})$ and ${v_j} = ({T_{j1}} \to {T_{j2}})$ are caused by close tracks crossing each other, a reasonable assumption is that if ${T_{i1}}$ is associated to ${T_{i2}}$, then there is a high probability that ${T_{j1}}$is associated to ${T_{j2}}$ and vice versa, if ${T_{i1}}$ is not associated to ${T_{i2}}$, there is a high probability that ${T_{j1}}$ should not be associated to ${T_{j2}}$.

Another kind of difficult node pairs come from the physical constraint that allows one hypothesis to be explained by at most one target. We apply repellency constraints in this case. An illustration is shown in Fig.~\ref{fig:Fig3}(b), where tracklet ${T_{i1}} = {T_{j1}}$ is shared by node ${v_i}$ and ${v_j}$. There exists a mutual exclusion to the labeling since two objects cannot occupy the same space at the same time. Specifically, labels of ${v_i}$  and ${v_i}$ cannot be set to ``1'' simultaneously due to time overlapping of the two nodes. It is worth noting that labeling repellency is helpful to alleviate the issue of assigning one tracklets to more than one targets.
\begin{figure}
\begin{center}
\includegraphics[width=82mm]{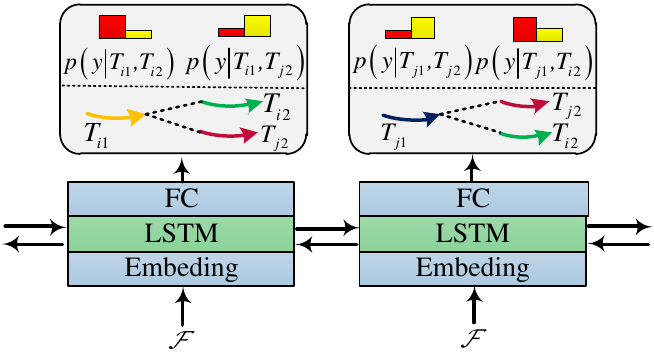}
\end{center}
\vspace{-4mm}
\caption{Two time step LSTM model for labeling difficult node pairs. At each time step, the network outputs a probability distribution for the joint matching problem, i.e., one element from set ${\cal A}$ to each element in set ${\cal B}$. Here $y \in \{ 0,1\}$ is a binary decision variable representing an assignment between two tracklets.}
\label{fig:Fig4}
\vspace{-4mm}
\end{figure}

\vspace{-3mm}
\paragraph{Potential Functions.} So far, we have introduced labeling consistency/repellency constraints for difficult node pairs, which encode dependencies among targets and could be exploited for CRF inference. Different from existing MOT methods that are limited by hand-designed pairwise potentials, the key idea of our solution is to avoid the need to specify the explicit knowledge among difficult node pairs. Instead, we use highly expressive neural networks to learn their functions directly from the data.

In this paper, we formulate labeling of difficult node pairs as a joint matching problem. As illustrated in Fig.~\ref{fig:Fig4}, a bidirectional LSTM-based architecture is learned for this label assignment.

 For a node pair $\{ {v_i}({T_{i1}} \to {T_{i2}}),{v_j}({T_{j1}} \to {T_{j2}})\}$, we first build two tracklet sets ${\cal A} = \{ {T_{i1}},{T_{j1}}\}$ and ${\cal B} = \{ {T_{i2}},{T_{j2}}\}$ as our matching graph. The goal here is to find correspondences between tracklets in the two sets. Note that we do not impose constraints to maintain a one-to-one relationship between ${\cal A}$ and ${\cal B}$, i.e., points in ${\cal A}$ can be assigned to more than one points in ${\cal B}$.

The input feature ${\cal F}$  that connects visual and motion cues is first passed through an embedding layer and then fed into a two time step bidirectional LSTM~\cite{Hochreiter1997Long}. %See more details about the used deep features in the supplementary document.
The network sequentially outputs a joint probability distribution for matching problem, i.e. $p\left( {y|{T_{i1}},{T_{i2}}} \right)$, $p\left( {y|{T_{i1}},{T_{j2}}} \right)$, $p\left( {y|{T_{j1}},{T_{i2}}} \right)$ and $p\left( {y|{T_{j1}},{T_{j2}}} \right)$. Here $y \in \{ 0,1\}$ is a binary decision variable representing an assignment between two tracklets. Finally, we define pairwise potential function for the difficult node pair $\{ {v_i},{v_j}\}$ as
\begin{equation}
{\varphi _{i,j}}({x_i},{x_j}) =  - {\omega _d}log({z_{i:{x_i}}} \cdot {z_{j:{x_j}}} + \varepsilon )
\label{Eq:eq4}
\end{equation}
where ${z_{i:{x_i}}} = p\left( {{x_i}|{T_{i1}},{T_{i2}}} \right)$ and ${z_{j:{x_j}}} = p\left( {{x_j}|{T_{j1}},{T_{j2}}} \right)$ represent the probabilities of joint matching state over $v_i$ and $v_j$. ${\omega _d}$ is a parameter controlling the impact of potentials and $\varepsilon$ is introduced to avoid numerical problems for small values of ${z_{i:{x_i}}} \cdot {z_{j:{x_j}}}$.

\begin{figure}[t]
	\begin{center}
		\includegraphics[width=80mm]{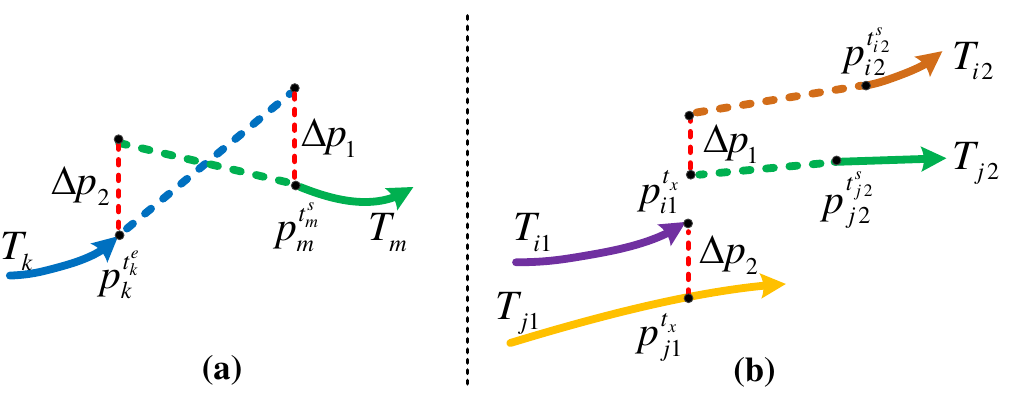}
	\end{center}
	\caption{An illustration of the motion vectors between two tracklets and node pairs.}
	\label{fig:Fig5s}
\end{figure}

%%%%%%%%%%%%%%%%%%%%%%%%%%%%%%%%xjxjxjxjxjxjxj%%%%%%%%%%%%%%%%%%%%%%%%%%%%%%%%%%%%%%%%%
\vspace{-3mm}
\paragraph{Deep feature ${\cal F}$.}The deep feature ${\cal F}$ used to represent pairwise potential consist of appearance and motion cues over node pairs as follows:
\begin{equation}
\label{eq:s1}
\begin{array}{c}
{\cal F}\! =\! [{f_a}\!\left( {{T_{i1}}} \right)\!,\!{f_a}\!\left( {{T_{i2}}} \right)\!,\!{f_a}\!\left( {{T_{j1}}} \right)\!,\!{f_a}\!\left( {{T_{j2}}} \right)\!,\!{f_m}\!\left( {{T_{i1}}\!,\!{T_{i2}}} \right)\!,\!\\
{f_m}\!\left( {{T_{i1}}\!,\!{T_{j2}}} \right)\!,\!{f_m}\!\left( {{T_{j1}}\!,\!{T_{i2}}} \right)\!,\!{f_m}\!\left( {{T_{j1}}\!,\!{T_{j2}}} \right)\!,\!{f_m}\!({v_i}\!,\!{v_j})]
\end{array}
\end{equation}
In Eq. \ref{eq:s1}, the deep appearance feature ${f_a}\left( {{T_i}} \right)$ is generated by inputting the most confident detection in $T_i$ to a Re-ID network (see Section 4.2 for details). ${f_m}\left( {{T_{k}},{T_{m}}} \right)$ means the motion feature between the tracklet $T_{k}$ and $T_{m}$, defined by the distance between estimations of positions of two tracklets using a linear motion model and the real positions as in Kuo et al.~\cite{discriminativeAppKuo2010MultitargetTB} and Yang et al.~\cite{Bo2014Multi}. As shown in Fig.~\ref{fig:Fig5s} (a), motion vector of $T_k$ and $T_m$ is defined as ${f_m}({T_k},{T_m}) = [\Delta {p_1},\Delta {p_2}]$, where distances are computed as $\Delta {p_1} = p_k^{t_k^e} + {v_k^{t_k^e}} \cdot \left( {t_m^s - t_k^e} \right) - p_m^{t_m^s}$ and $\Delta {p_2} = p_m^{t_m^s} - {v_m^{t_m^s}} \cdot \left( {t_m^s - t_k^e} \right){\rm{ - }}p_k^{t_k^e}$. Here $v_{k}^{t}$ stands for the velocity of tracklet ${T_{i}}$ at time step $t$. %Here $v_k^{t_k^e}$ and $v_m^{t_m^s}$ are the velocity vectors for tracklet ${T_k}$ and ${T_v}$ at time step ${t_k^e}$ and  ${t_m^s}$ respectively.
 Finally, ${f_m}({v_i},{v_j})$ is a distance function for a difficult node pair. As shown in Fig.~\ref{fig:Fig5s} (b), considering node pair $\{ {v_i}({T_{i1}} \to {T_{i2}}),{v_j}({T_{j1}} \to {T_{j2}})\}$ and supposing $T_{i2}$ and $T_{j2}$ are a head-close tracklet pair. Let ${t_x} = \min (t_{i1}^e,t_{j1}^e)$, we estimate positions of both targets at frame $t_x$, shown by black points in Fig.~\ref{fig:Fig5s} (b). The distance of the node pair is defined by the estimated relative distances and real position distances at frame $t_x$, formulated as ${f_m}({v_i},{v_j}) = [\Delta {p_1},\Delta {p_2}]$, where relative distances $\Delta {p_1} = p_{i2}^{t_{i2}^s} - v_{i2}^{t_{i2}^s} \cdot (t_{i2}^s - {t_x}) - (p_{j2}^{t_{j2}^s} - v_{j2}^{t_{j2}^s} \cdot (t_{j2}^s - {t_x}))$ and $\Delta {p_2} = p_{i1}^{{t_x}} - p_{j1}^{{t_x}}$.

	\subsection{Inference as RNN}
One challenge in CRF inference is that a realization of assignment hypothesis takes values from a discrete set, e.g., \{0, 1\}. However, existing deep learning methods are not designed for the discrete problem. To address this issue and use DNNs to produce optimal assignment hypothesis, we first take continuous relaxation of the original binary labelling and then pose it as an optimization problem like in~\cite{KShortestBerclaz2011MultipleOT,ArbitraryLarsson2017LearningAP}. Specifically, the original minimization for Eq.~\ref{Eq:eq2} is first transformed into an equivalent integer program problem by expanding label variable $x_i$ into another new Boolean variables ${x_{i:\lambda }},\lambda  \in \{ 0,1\}$. ${x_{i:\lambda }}$ represents an assignment of label $\lambda$ to $x_i$, which is equivalent to an assignment of Boolean labels 0 or 1 to each node ${x_{i:\lambda }}$, and an assignment of label 1 to ${x_{i:\lambda }}$ means that $x_i$ receives label $\lambda$. A constraint is introduced to ensure that only one label value is assigned to each node, i.e, $\sum\limits_{\lambda  \in {\cal L}} {{x_{i:\lambda }}{\rm{ = 1}}}$. In this way, we rewrite the original energy minimization as the following binary integer program
	\begin{equation}
	\begin{array}{l}
	\min \sum\limits_{i \in {\cal V},\lambda  \in {\cal L}} {{\varphi _i}(\lambda )\,} {x_{i:\lambda }}\! + \!\sum\limits_{\scriptstyle(i,j) \in {\cal E}\hfill\atop
		\scriptstyle\lambda ,\mu  \in {\cal L}\hfill} {{\varphi _{ij}}(\lambda ,\mu )\,{x_{i:\lambda }}{x_{j:\mu }}} \\
	s.t.\;\;\quad {x_{i:\lambda }} \in \{ 0,1\} \quad \quad \quad \forall i \in {\cal V},\,\lambda  \in {\cal L}\\
	\quad \quad \;\;\sum\limits_{\lambda  \in {\cal L}} {{x_{i:\lambda }} = 1} \quad \quad \quad \;\;\forall i \in {\cal V}
	\end{array}
    \label{Eq:eq5}
	\end{equation}
	
	As a next step, we relax the integer program by allowing real values on the unit interval [0, 1] instead of Booleans only. Let ${q_{i:\lambda }} \in [0,1]$ denote the relaxed variables, and energy functions can be represented as the following quadratic program
	\begin{equation}
	\begin{array}{l}
	\min \sum\limits_{i \in {\cal V},\lambda  \in {\cal L}} {{\varphi _i}(\lambda )\,} {q_{i:\lambda }}\! + \!\sum\limits_{\scriptstyle(i,j) \in {\cal E}\hfill\atop
		\scriptstyle\lambda ,\mu  \in {\cal L}\hfill} {{\varphi _{ij}}(\lambda ,\mu )\,{q_{i:\lambda }}{q_{j:\mu }}} \\
	s.t.\;\;\quad {q_{i:\lambda }} \in [0,1]\quad \quad \quad \forall i \in {\cal V},\,\lambda  \in {\cal L}\\
	\quad \quad \;\;\sum\limits_{\lambda  \in {\cal L}} {{q_{i:\lambda }} = 1} \quad \quad \quad \;\;\forall i \in {\cal V}
	\end{array}
    \label{Eq:eq6}
	\end{equation}
%Inspired by the previous work~\cite{ArbitraryLarsson2017LearningAP,CRFasRNNZheng2015ConditionalRF},
	 By now, we can take the CRF inference in Eq.~\ref{Eq:eq6} as a gradient descent based minimization problem which can easily be formulated in a recurrent neural network manner, since all of the operations are differentiable with respect to $\mathbf{q}$. Next, we  describe how to implement one iteration of gradient descent  for our CRF inference with common operations of neural network layers.
	
	\vspace{-3mm}
	\paragraph{One Iteration of CRF Inference.}
	We use the standard gradient method to minimize the energy function in Eq.~\ref{Eq:eq6}, The gradient of the objective function in Eq.~\ref{Eq:eq6}, ${\nabla _q}E$, has the following elements
	
	\begin{equation}
	\frac{{\partial E}}{{\partial {q_{i:\lambda }}}} = {\varphi _i}(\lambda ) + \sum\limits_{j: \scriptstyle(i,j) \in {\cal E}\hfill\atop
		\scriptstyle\mu  \in {\cal L}\hfill} {{\varphi _{ij}}(\lambda ,\mu )\,{q_{j:\mu }}}
    \label{Eq:eq7}
	\end{equation}
	We denote the contribution from pairwise term by
	\begin{equation}
	{q'_{i:\lambda }} = \sum\limits_{j: \scriptstyle(i,j) \in {\cal E}\hfill\atop
		\scriptstyle\mu  \in {\cal L}\hfill} {{\varphi _{ij}}(\lambda ,\mu )\,{q_{j:\mu }}}
    \label{Eq:eq8}
	\end{equation}
	and decompose ${q'_{i:\lambda }}$ into the following two operations which can be described as CNN layers
	
	\begin{equation}
	{q'_{i:\lambda }}(\mu ) = \sum\limits_{j:(i,j) \in{\cal E} } {{\varphi _{ij}}(\lambda ,\mu )}  \cdot {q_{j:\mu }}
    \label{Eq:eq9}
	\end{equation}
	\begin{equation}
	{q'_{i:\lambda }} = \sum\limits_{\mu  \in {\cal L}} {{{q'}_{i:\lambda }}(\mu )}
    \label{Eq:eq10}
	\end{equation}
	
	In Eq.~\ref{Eq:eq9}, ${q'_{i:\lambda }}(\mu )$ is obtained by applying a filter on ${q_{j:\mu }}$, which can be implemented through a stand convolution layer. Here the filter's receptive field spans the whole nodes and coefficients of the filter are derived on the pairwise potential functions ${\varphi _{ij}}(\lambda ,\mu )$ . For ${q'_{i:\lambda }}$ in Eq.~\ref{Eq:eq10}, the operation step is to take a sum of the filter outputs for each class. Since each class label is considered individually, this can be viewed as usual convolution where the spatial receptive field of the filter is $1 \times 1$, and the number of input or output channels is ${\rm{|}}{\cal L}{\rm{|}}$. To get the complete gradient in Eq.~\ref{Eq:eq7}, the output from the pairwise stage and the unary term ${\varphi _i}(\lambda )$ are combined using an element-wise summing operation. Then according to the gradient update principle, the new state of $\mathbf{q}$ can be computed by taking a step in the negative direction of gradient:
	\begin{equation}
	{\tilde q^{t + 1}} = {q^t} - \gamma \;{\nabla _q}E
    \label{Eq:eq11}
	\end{equation}
	where $\gamma$ is the step size. Finally, we normalize our values to satisfy $\sum\limits_{\lambda  \in {\cal L}} {{q_{i:\lambda }} = 1}$ and  $0 \le {q_{i:\lambda }} \le 1$ by applying a Softmax operation with no parameters.
	
\begin{figure}
		\begin{center}
			\includegraphics[width=82mm]{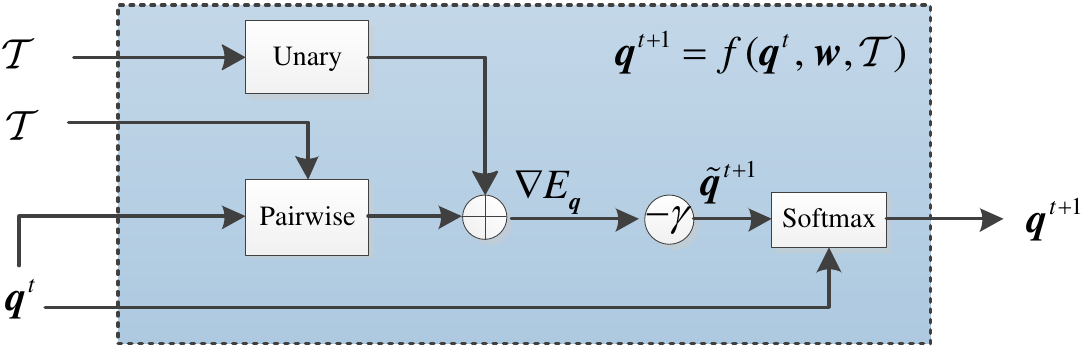}
		\end{center}
        \vspace{-3mm}
		\caption{Data flow of one iteration of the gradient descent algorithm. Each rectangle or circle represents an operation that can be performed as the standard convolutional operations.}
		\label{fig:Fig7}
    \vspace{-3mm}
	\end{figure}

	\vspace{-3mm}
	\paragraph{Integration in A Recurrent Neural Network.}
	We have formulated one iteration of the CRF inference into a stack of common CNN operations. Now we can organize the entire CRF inference as RNN architecture, by unrolling the iterative gradient descent algorithm as a recurrent neural network and all the CRF's parameters can be learned or fine-turned end-to-end via back propagation. The data flow of one iteration of RNN inference is shown in Fig. \ref{fig:Fig7}.  The RNN takes tracklets' observations ${\cal T}$ for both unary and pairwise probability as input. Each iteration of RNN in the forward pass performs one step of gradient descent for CRF inference, denoted by ${\mathbf{q}^{t + 1}} = f({\mathbf{q}^t},\;w,\,{\cal T})$, where $w$ represents a set of parameters containing ${\omega _u}$, ${\omega _p}$, $\gamma$ as well as filter weights for pairwise and unary terms. In the stage of initialization (i.e., at the first time step), ${\mathbf{q}^0}$ is set as the probability output for unary. At all other time steps, ${\mathbf{q}^t}$ is set to be ${\mathbf{q}^{t+1}}$. RNN outputs nothing but ${\mathbf{q}^T}$ until the last iteration that approximates a solution to Eq.~\ref{Eq:eq6}
	
	\subsection{Final deep structure model}
	The entire structure of our framework is shown in Fig.~\ref{fig:Fig2}. Given graph nodes, the first part of our model consists of a unary and pairwise architecture to produce potential values of nodes. The second part is a RNN model that performs CRF inference with gradient descent algorithm. Since all the pieces of the model are formulated into standard network operations, our final model can be trained in the manner of end-to-end using common deep learning strategies. To be specific, during forward pass, the input nodes are fed into CNNs and LSTM to compute unary and pairwise potentials. The output by CNN is then passed to RNN as initialization, i.e. CRF state $q^0$. After $T$ iterations, the RNN finally outputs $q^T$ , the solution to Eq.6. During backward pass, the error derivatives are first passed through the RNN, where the gradients w.r.t. the parameters and input accumulate through the $T$ iterations. The gradients are back propagated to unary and pairwise components, and then update is performed for the CNN and LSTM simultaneously. In other word, all the CRF learning and inference are embedded within a unified neural network.
	\section{Implementation Details}
	
	This section describes the details of the proposed method in both the training and testing stages. %In our opinion, it is infeasible to train the whole model from scratch due to the limited training trajectories and vanishing/exploding gradient problems for RNN.
	In this work, we finetune the VGG-16 model~\cite{Large-Scale2014VeryDC} by the Re-ID training sets as feature extractor and learn LSTM from scratch. The feature extractor and LSTM modules are updated within the RNN framework. %Note that VGG-16 can be replaced by other common CNN architectures.

	\subsection{Hyperparameter Choices}
	The proposed CRF model has two learnable parameters.

\begin{figure}
		\begin{center}
			\includegraphics[width=80mm]{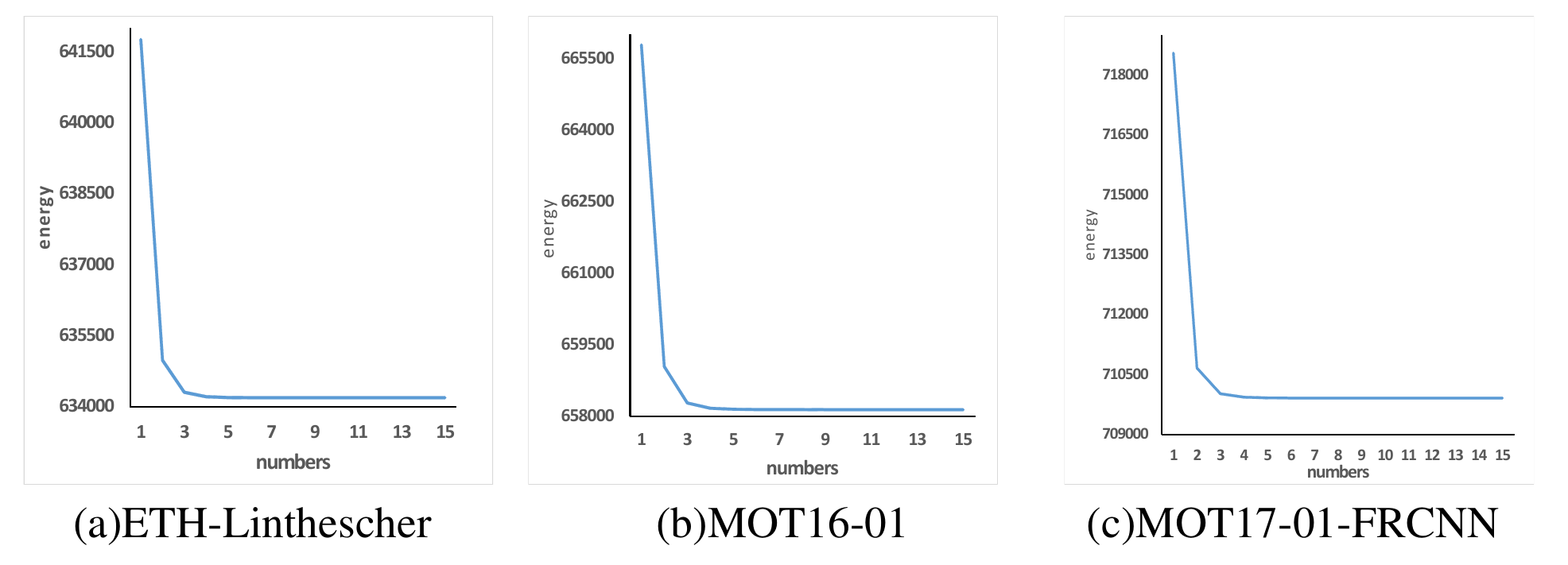}
		\end{center}
		\vspace{-3mm}
		\caption{Illustration of the energy decreasing over training iterations on three datasets. Horizontal axis is the iteration number and vertical axis is the energy value.}
		\vspace{-3mm}
		\label{fig:Fig10}
	\end{figure}

The first is the number of iterations $T$ in the gradient decent algorithm. In our experiments, we set the number of $T$ to 5 during the training stage to avoid the gradient  vanishing problem. %Experimental, we add a test to show whether the energy of Eq.7 decreases over time.
	Fig.~\ref{fig:Fig10} shows the energy of Eq. \ref{Eq:eq7} converges fast (after 5 iterations) on three test datasets.

The second parameter is ${T_{thr}}$, the maximal time interval of linkable tracklet pairs to build the graph. Intuitively, a small threshold ${T_{thr}}$ would be inefficient to deal with long-term missing detections, hence resulting in more tracking fragments. In experiments, we adopt two-level association to gradually link tracklets into final tracks. For all dataset, ${T_{thr}}$ is empirically set to 20. The lower the threshold is, less likely are the short tracklets to be associated, leading to fewer ID switches but more fragments. In fact, missing detections, false detections and unassociated detections inevitably lead to fragments during the first round of association. To alleviate this problem, a second round association with ${T_{thr}=50}$ is performed over the output of the first round to reconnect fragment tracks.
	
	\subsection{Architecture and Data Collection}
	%It is more desirable to use detection bounding boxes for model training, since the characteristics of detections and ground-truths are different and only detection bounding boxes are available for inference.
	We train our models using detection bounding boxes with associated IDs from ground-truth labels and randomly shift the center, width and height of the boxes. To avoid confusion, only detections whose visibility scores are larger than 0.5 are selected. Inspired by~\cite{Discrete-Continuous2016MultiTargetTB,HFRFXiang2016MultitargetTU}, we design an explicit occlusion reasoning model to compute the visibility for each target.
	
	\vspace{-3mm}
	\paragraph{Unary Potentials.} Our base CNN architecture is the VGG-16 model~\cite{Large-Scale2014VeryDC}, denoted as ID-Net. We collect training set from two different datasets. We collect training images firstly from the 2DMOT2015 benchmark training set~\cite{leal2015motchallenge} and 5 sequences from the MOT16 benchmark training set~\cite{milan2016mot16}. We also collect person identity examples from the CUHK03~\cite{CUHK03Zheng2015ScalablePR} and Market-1501~\cite{hermans2017defense} datasets. For validation set, we use the MOT16-02 and MOT16-11 sequences from the MOT16 training set. In total, 2551 identities are used for training and 123 identities for validating. Specifically, we train VGG-16 from scratch with $Y = 2551$  unique identities. % the learning can be viewed as a  $Y$-way classification problem.
	Training images are re-sized to $112 \times 224 \times 3$ and each image ${I_i}$ corresponds to a ground-truth identity label ${l_i} \in \{ 1,2, \cdots Y\}$. The network is trained with the softmax loss. % to estimate the probability of each image being each label by a forward pass.
	A Siamese network is learned by fine-turning the ID-Net with a binary softmax function. % to produce the probability of a image pair belonging to the same target or not.
	We set the batch size to 64, momentum to 0.9, dropout ratio for FC layers to 0.5. The learning rate is initially set to 0.01, and then decreased by a factor of 10 when accuracy in the validation set stopped improving. The maximum number of iterations is set to 40000.
	
	\vspace{-3mm}
	\paragraph{Pairwise Potentials.} To learn visual features, we directly employ a pre-trained person Re-Identification network~\cite{DefenseTripHermans2017InDO}, which is the state-of-the-art Re-ID network using a variant of triplet loss, and then fine-tune it on our MOT datasets. We collect triplet examples from the MOT15 benchmark training set and 5 sequences of the MOT16 benchmark training set. The MOT16-02, MOT16-09 sequences in the MOT-16 training set are used as testing sets. Overall 888 identities are used for training and 79 identities for validation. Triplet examples are generated as follows: for each batch of 100 instances, we select 5 persons and generate 20 instances for each person. In each triplet instance, the anchor and anchor positive are randomly selected from the same identity, and the negative one is also randomly selected, but from the remaining identities. The triplet loss margin is set to -1. Here the final deep feature ${\cal F}$ to represent each node pair is a $4 \times 128 + 10 \times 2 = 532$ dimensional vector .
	
	For the bidirectional LSTM, an FC layer is employed to map the entire input ${\cal F}$ into a 300 dimensional embedding space. We use a two-layer hidden state vector of size 200 and a FC layer with dropout 0.5 to produce the $1 \times 2 \times 2$ distribution output at each step. During training, 65000 difficult node pairs are collected by randomly sampling tracklets from true target trajectories. We divide this data into mini-batches where each batch has 32 samples. We normalize position data to the range [-0.5, 0.5] with respect to image dimensions. We use the Adam algorithm~\cite{AdamOpKingma2014AdamAM} to minimize the cross-entropy loss. The learning rate is set to 0.0001 and the maximum number of iterations is set to 40000 for convergence.
	
	\vspace{-3mm}
	\paragraph{CRF Inference.} We generate tracklets by randomly sampling 6 sequences of the MOT16 dataset as training data and use the MOT16-02 sequence as the validation set. The learning parameters of the RNN are the impact coefficients ${w_u},{w_d}$, and learning step $\gamma$. We fine-tune unary and pairwise potentials simultaneously. To handle the issue of varying number of targets in learning neural network architectures, we adopt the sliding window strategy to produce a fixed number of nodes, then shift the window forward to ensure that the new sliding window overlaps with the previous region. This allows tracklets to be linked over time. Considering the trade-off between accuracy and efficiency, the node number in sliding window is set to 200. We use the Adam algorithm~\cite{AdamOpKingma2014AdamAM} to minimize the cross-entropy loss with 200 nodes per batch. The learning rate is set to 0.001. ${w_u},{w_d}$ are all initialized to 1, and the initial learning step $\gamma$ is set to 0.5. The maximum number of iterations is set to 60000.
	
	%%%%%%%%%%%%%%%%%%%%%%%%%%%%%%%%%%%%%%%%%%%%%%%%%%%%%%%%%%%%%%%%%%%%%%%%%%%%%%%%%%%%%%%%%%%%%%%%%%%%%%%
	\section{Experimental Results }
	In this section, we first describe the used evaluation metrics. Then we present ablation studies on each component of our CRF framework. Finally, we report the overall performance in comparison with state-of-the-art approaches on  three MOT benchmark datasets.
	
	\subsection{Evaluation Metrics}
	We follow the standard MOT2D Benchmark challenge~\cite{leal2015motchallenge,milan2016mot16} for evaluating multi-object tracking performance. These metrics include: Multiple Object Tracking Accuracy (MOTA $\uparrow$),
The Ratio of Correctly Identified Detections(IDF1 $\uparrow$),Multiple Object Tracking Precision (MOTP $\uparrow$), Mostly Tracked targets (MT $\uparrow$), Mostly Lost targets (ML $\downarrow$), False Positives (FP $\downarrow$), False Negatives (FN $\downarrow$), Fragmentation (FM $\downarrow$), ID Switches (IDS $\downarrow$),finally Processing speed(HZ $\uparrow$). The up arrow $\uparrow$ denotes that the higher value is better and the down arrow $\downarrow$ represents the opposite. %, and definition statement of the metrics are  detailed in  Tab ~\ref{tab:table7}
	
	\subsection{Ablation Studies}
	We follow the standard CRF based MOT framework to analyze the contributions of different components. Specifically, we investigate the contribution of different components in our CRF framework with detailed tracking metrics as well as visualized results on the MOT16-02 and MOT16-09 datasets by separately studying unary terms only, unary + pairwise, as well as different inference strategies. To this end, we evaluate three alternative implementations of our algorithm with the same deep architectures as follows:
	\begin{itemize}
		\itemsep 0mm
		\item Unary terms only (U). We use the Hungarian algorithm to find the globally optimal solution. In this case, we only consider appearance information of targets.
		\item CRF with an approximate solution in polynomial time (U+P). We use unary and pairwise terms but without global RNN inference. We replace the potentials in~\cite{Bo2014Multi} with ours and conduct CRF inference with  the heuristic algorithm~\cite{Bo2014Multi}, which suggests that our potentials work in the conventional  optimization algorithm.
		\item The proposed method (CRF-RNN) equipped with all the components.
	\end{itemize}
	
%	``U'' means tracking method using . In this case, we use the Hungarian algorithm to find a global optimal solution. Note that in such situation, it only considers targets ��?appearance information.
%	
%	``U+P'' means the introduced heuristic algorithm in~\cite{Bo2014Multi} to find an approximate solution of CRF in polynomial time. Specifically, such tracking method used unary and pairwise terms but without global RNN inference. It is interesting that we design algorithm ``U+P'' by replacing the potentials in~\cite{Bo2014Multi} with ours and conducting CRF inference with heuristic algorithm in ~\cite{Bo2014Multi},which suggests that our potentials also work in a regular optimization algorithm.
%	
%	``CRF-RNN'' represents the proposed model.
	
	Table~\ref{tab:table1} shows that unary terms are effective for tracking, meaning that the appearance cue is important for MOT. %In fact people appearance can be learned for varying viewpoint and motion, which is less easy to achieve by motion models.
	The pairwise terms help to improve performance, especially in highly crowded scenes with clutter and occlusions. Comparing the U+P method with the U method, MOTA is increased from 25.3\% to 26.7\%. %,  which indicates the effectiveness of paiwise terms.
	By using pairwise terms, spatial-temporal dependencies can be effectively modeled to distinguish close targets.

\begin{figure}
		\begin{center}
			\includegraphics[width=80mm]{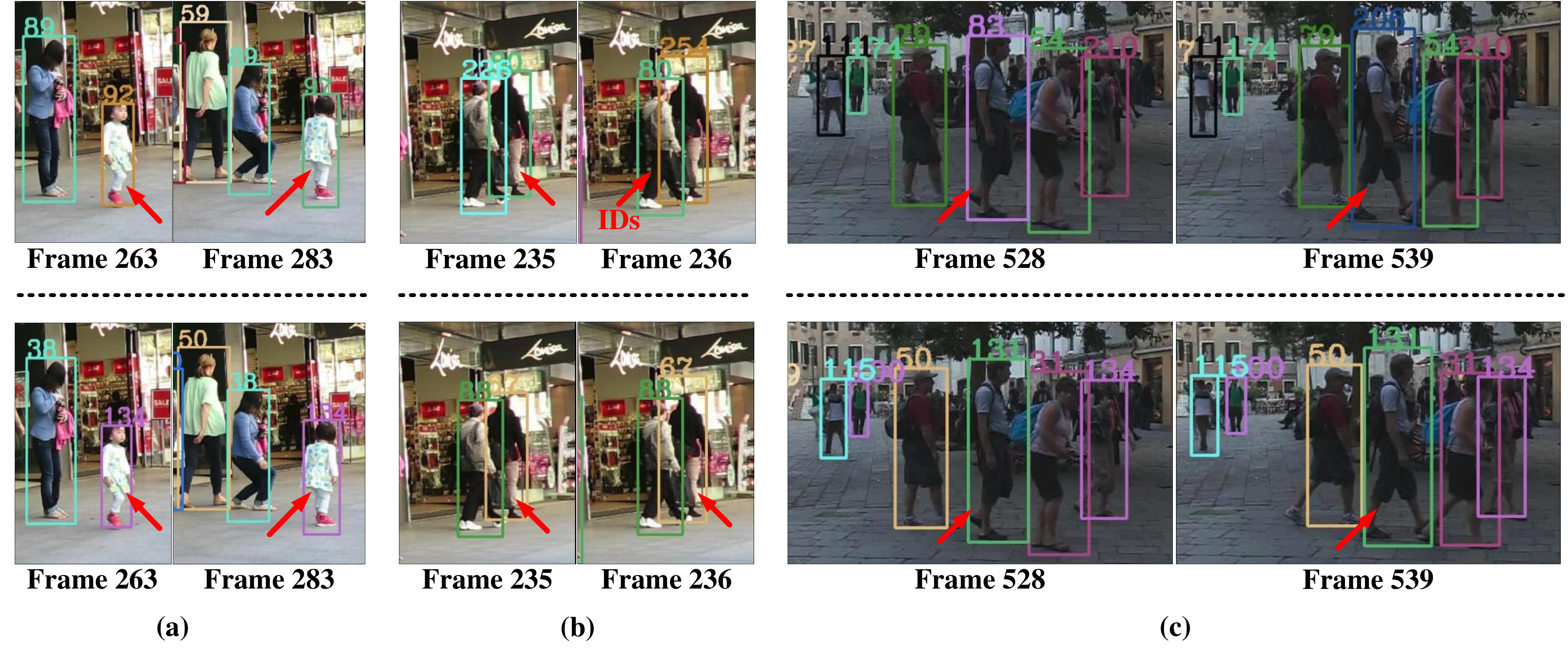}
		\end{center}
		\vspace{-4mm}
		\caption{Qualitative comparison. The first row shows tracking results  using unary terms only (U). The second row shows tracking results considering both unary and pairwise terms (U+P).}
		\label{fig:Fig8}
		\vspace{-1mm}
	\end{figure}

Fig.~\ref{fig:Fig8} visually compares the results of these two implementations.
	%In the first row of Figure ~\ref{fig:Fig8}(a), a fragment occurs at frame 283 when a little girl in write makes a sharp direction change which leads to failure of the appearance model. In the second row of Figure ~\ref{fig:Fig8}(a), by incorporating pairwise terms, we are able to locate the non-linear path and to track the girl correctly.
    In the first row of Fig.~\ref{fig:Fig8}(a), a fragment occurs at frame 283 when a little girl in write makes a sharp direction change which leads to failure of the appearance model. In the second row of Fig.~\ref{fig:Fig8}(a), by incorporating pairwise terms, we are able to locate the non-linear path and to track the girl correctly.

	In the first row of Fig.~\ref{fig:Fig8}(b), an ID switch occurs between two close persons due to occlusions. However, by using pairwise terms, the U+P method can distinguish the targets, as shown in the second row of Fig.~\ref{fig:Fig8}(b).

Fig.~\ref{fig:Fig8}(c) indicates that the pairwise model is also robust to illumination variations.

\begin{figure}
		\begin{center}
			\includegraphics[width=80mm]{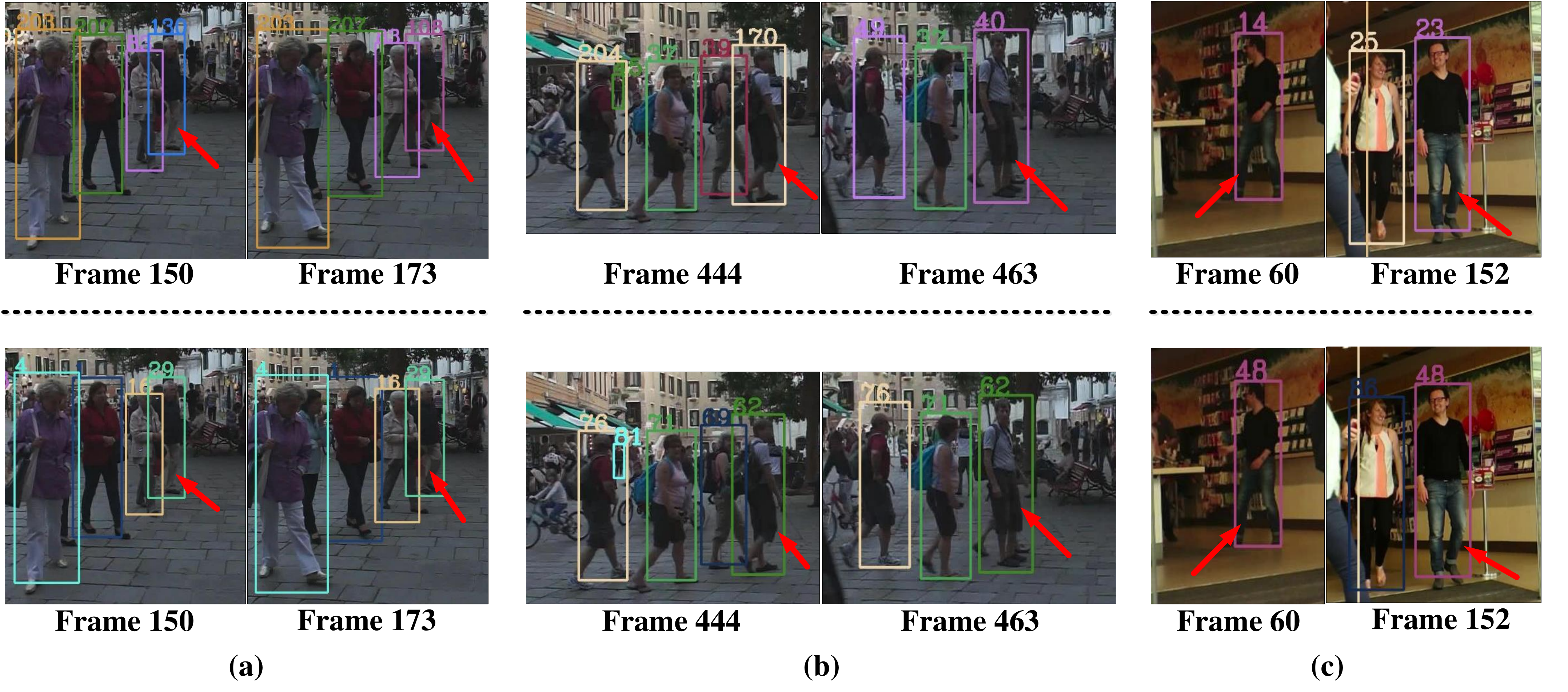}
		\end{center}
		\vspace{-4mm}
		\caption{Qualitative comparison. The first row shows tracking results of of considering both unary and pairwise terms with the regular optimization algorithm (U+P). The second row shows the tracking results of our deep CRF method.}
		\vspace{-1mm}
		\label{fig:Fig9}
	\end{figure}

	The proposed CRF-RNN method outperforms both the U and U+P baselines in all metrics. This clearly shows that during CRF inference both unary and pairwise cues are complementary to contribute to tracking performance improvement. Fig.~\ref{fig:Fig9} also shows that our method is more robust to heavy occlusion, crowded scenes and illumination varying.
	
	We also conducted an experiment by fixing the unary and pairwise parts and learning only the RNN component (i.e., the CRF inference model). In this case, the performance of MOTA is increased by only 0.5\% compared to the U+P baseline. Note that MOTA is the most important metric to evaluate tracking performance. Considering that the proposed deep CRF method achieves the improvement of 2.1\%  over the U+P method, it is clear that end to-end training is helpful to boost the tracking accuracy significantly.
	
	%\begin{figure}
	%\begin{center}
	%\includegraphics[width=80mm]{Figure8.pdf}
	%\end{center}
	%\caption{Comparison of visual results. The first row shows tracking results of ``U+P'' (i.e., considering both unary and pairwise terms). The second row shows tracking results of our proposed CRF-RNN method.}
	%\label{fig:Fig8}
	%\end{figure}
	
	%-------------------------------------------------------------------------
	
	\begin{table}[tbp]\footnotesize
		\caption{Ablation studies on the MOT16 validation set.}
		\vspace{-2mm}
		\label{tab:table1}
		\begin{center}
			\resizebox{.48\textwidth}{!}{
			\begin{tabular}{|p{13mm}|p{8mm}p{8mm}p{8mm}p{8mm}p{8mm}p{8mm}p{8mm}|}
				\hline
				Tracker & MOTA$\uparrow$ & MOTP$\uparrow$ & MT$\uparrow$ & ML$\downarrow$ & FP$\downarrow$ & FN$\downarrow$ & IDS $\downarrow$\\
				\hline
				U & 25.3 & 74.9 & \textbf{11.4\%} & 50.6\% & 600 & 16294 & 359\\
				U+P & 26.7 & 75.0 & \textbf{11.4\%} & 50.6\% & 523 & 16118 & 281 \\
				CRF-RNN & \textbf{28.8} & \textbf{75.1} & \textbf{11.4\%} & \textbf{48.1\%} & \textbf{317} & \textbf{15943} & \textbf{173} \\
				\hline
			\end{tabular}
		}
		\end{center}
		\vspace{-2mm}
	\end{table}
	
	\subsection{Comparison with the State of the Art}
	
    We compare our CRF-RNN method with the top published works on the MOT16 test set, and report the quantitative results in Table~\ref{tab:table8}. Our method achieves new state-of-the-art performance in terms of MOTA, ML and FN, and ranks the 2nd with MT of 18.3. The higher MT and lower ML suggest that our method is capable of recovering targets from occlusion or drifting. The recent work ~\cite{DCCRFZhou2018Deep} proposes a deep continuous Conditional Random Field (DCCRF) framework to solve the MOT problem. Our method performs better than DCCRF in most metrics, such as MOTA (50.3\% versus 44.8\%), IDF1(54.5\% versus 39.7\%), MT (18.3\% versus 14.1\%), ML (35.7\% versus 42.3\%) and IDS (702 versus 968). Our method also has significant advantages over LTCRF~\cite{CRF-BasedLe2016LongTermTC}, another CRF-based approach.

    The main reason for high FP is that our method uses linear interpolation to connect fragments, which is unable to produce more accurate prediction in complex scenes. When targets are heavily occluded, the proposed method fails to assign them with correct tracklets, leading to a relative more number of switches. Nevertheless, we have a higher MT score. By incorporating more effective occlusion reasoning strategies to compute the similarity between tracklets, our results can be further improved.

    We also conduct experiment on 2DMOT2015 dataset and report the results in Table~\ref{tab:table9}. Similar to above discussion, our method achieves the best metrics in terms of MOTA (40\%), IDF1(49.6\%), and FN (25917). We obtain the second best record in terms of MT(23.0\%) and ML(28.6\%).

	%%%%%%%%%%%%%%%%%%%%%%%%%%%%%%%%%%%%%%%%%%%%%%%%%%%%%%%%%%
\begin{table*}
\caption{Results on the MOT16 test dataset. The best and second best results are highlighted in bold and red.}
\label{tab:table8}
\begin{center}
\begin{tabular}{|l|cccccccccc|}
\hline
Tracker & Mode & MOTA$\uparrow$ & IDF1$\uparrow$ & MOTP$\uparrow$ & MT$\uparrow$ & ML$\downarrow$ & FP$\downarrow$ & FN$\downarrow$ & IDS $\downarrow$ & HZ$\uparrow$\\
\hline
\textbf{CRF-RNN(Ours)} & Batch &\textbf{50.3} &54.4& 74.8 & \color{red}18.3\% & \textbf{35.7\%} & 7148  & \textbf{82746} & 702 & 1.5 \\
        HCC~\cite{MaTangACCV2018}& Batch &\color{red}49.3 & 50.7 &\textbf{79.0} &17.8\% &39.9\% & \color{red}5333  & 86795 & \textbf{391} & 0.8 \\
		LMP~\cite{LiftedMulticutTang2017MultiplePT}& Batch &48.8 & 51.3 &\textbf{79.0} &18.2\% & 40.1\% & 6654  &86245 & 481 & 0.5 \\
        TLMHT~\cite{ShengIterative}& Batch &48.7 & \textbf{55.3} &76.4  &15.7\% & 44.5\% & 6632  & 86504 & \color{red}413 & \color{red}4.8 \\
        GCRA~\cite{Ma2018Trajectory}& Batch &48.2 & 48.6 &\color{red}77.5  &12.9\% & 41.1\% & 5104 & 88586 &821 & 2.8 \\
        eHAF16~\cite{Hao2018Heterogeneous}& Batch &47.2 & 52.4 &75.7  &\textbf{18.6\%} & 42.8 \% & 12586 & \color{red}83107 &542 & 0.5 \\
        NOMT~\cite{Near-OnlineChoi2015NearOnlineMT}& Batch &46.4 & 53.3 &76.6  &18.3\% & 41.1\% & 9753	& 87565	&359 & 2.6 \\
        Quad-CNN~\cite{QuadrupletSon2017MultiobjectTW} & Batch & 44.1 & 38.3 & 76.4 & 14.6\% & 44.9\% & 6388 & 94775 & 745 & 1.8 \\
        MHT-DAM~\cite{NHTKim2015MultipleHT} & Batch & 42.9 &47.8 & 76.6 & 13.6\% & 46.9\% & 5668 & 97919 & 499 & 0.8 \\
       % LINF1~\cite{fagot2016improving} & Batch & 41.00 &45.7	 & 74.80 & 11.60\% & 51.30\% & 7896 & 99224 & \textbf{430} & 1.1 \\
       LTCRF~\cite{CRF-BasedLe2016LongTermTC} & Batch & 37.6 & 42.1& 75.9 & 9.6\% & 55.2\% & 11969 & 101343 & 481 &0.6\\
\hline
        KCF16~\cite{Chu2019Online} & Online & 48.8 &47.2  & 75.7 & 15.8\% & \color{red}38.1\% & 5875  & 86567 & 906 & 0.1 \\
        AMIR~\cite{Sadeghian2017TrackingTU} & Online & 47.2 &46.3  & 75.8 & 14.0\% & 41.6\% & \textbf{2681}  & 92856 & 774 & 1.0 \\
        DMAN~\cite{Zhu2019Online} & Online & 46.1 &\color{red}54.8 & 73.8 & 17.4\% & 42.7\% & 7909	 & 89874	 & 532 & 0.3 \\
        DCCRF~\cite{DCCRFZhou2018Deep} & Online & 44.8 &39.7 & 75.6 & 14.1\% & 42.3\% & 5613 & 94125 & 968 &0.1\\
        CDA~\cite{ConfidenceBae2014RobustOM} & Online & 43.9 &45.1 & 74.7 & 10.7\% & 44.4\% & 6450 & 95175 & 676 & 0.5 \\
        %oICF~\cite{Kieritz2016Online} & Online & 43.2 & 74.30 & 11.30\% & 48.50\% & 6651 & 96515 & \textbf{381} & 0.4 \\
        EAMTT-pub~\cite{Sanchez2016Online} & Online & 38.8 &42.4 & 75.1  & 7.9\% & 49.1\% & 8114 & 102452 & 965 & \textbf{11.8} \\
        OVBT~\cite{Ban2016Tracking} & Online & 38.4 &37.8 & 75.4 & 7.5\% & 47.3\% & 11517 & 99463 & 1321 & 0.3 \\
\hline
\end{tabular}
\end{center}
\end{table*}

\begin{table*}
\caption{Results on the 2DMOT2015 test dataset. The best and second best results are highlighted in bold and red.}
\label{tab:table9}
\begin{center}
\begin{tabular}{|l|cccccccccc|}
\hline
Tracker & Mode & MOTA$\uparrow$ & IDF1$\uparrow$ & MOTP$\uparrow$ & MT$\uparrow$ & ML$\downarrow$ & FP$\downarrow$ & FN$\downarrow$ & IDS $\downarrow$ & HZ$\uparrow$\\
\hline
    \textbf{CRF-RNN(Ours)} & Batch & \textbf{40.0} & \textbf{49.6} & 71.9 & \color{red}23.0\% & {\color{red}28.6\%} & 10295 & \textbf{25917} & 658 & 3.2 \\
    JointMC~\cite{Keuper2018Motion} & Batch & 35.6 &45.1 & 71.9 & \textbf{23.2\%} & 39.3 \% & 10580	 & \color{red}28508	& 457  & 0.6 \\
    Quad-CNN~\cite{QuadrupletSon2017MultiobjectTW} & Batch & 33.8 &40.4 & \textbf{73.4} & 12.9\% & 36.90\% & 7898 & 32061 & 703 & 3.7 \\
    NOMT~\cite{Near-OnlineChoi2015NearOnlineMT} & Batch & 33.7 &44.6 & 71.9 & 12.2\% & 44.0\% & 7762 & 32547 & \color{red}442 & 11.5 \\
    MHT-DAM~\cite{NHTKim2015MultipleHT} & Batch & 32.4 &45.3 &71.8 & 16.0\% &43.8\% &9064 &32060 &\textbf{435} & 0.7\\
    CNNTCM~\cite{ConstrainedWang2016JointLO} & Batch & 29.6 &36.8 & 71.8 & 11.2\% & 44.0\% & 7786 & 34733 & 712 & 1.7\\
%LP-SSVM~\cite{LPSSVMWang2016LearningOP} & Batch & 25.20 &71.70 & 5.80\% &53.00\% &8369  &36932 &646 &41.3\\
    SiameseCNN~\cite{siamesecnn2016LearningBT} & Batch & 29.0 &34.3 &71.2 & 8.5\% &48.4\% & 5160 &37798 &639 &{\color{red}52.8}\\
\hline
        KCF~\cite{Chu2019Online} & Online & \color{red}38.9 &44.5  & 70.6 & 16.6\% & 31.5\% & 7321  & 29501 & 720 & 0.3 \\
        APHWDPLp~\cite{Long2018Online}& Online & 38.5 &47.1  & \color{red}72.6 & 8.7\% & 37.4\% & \textbf{4005}  & 33203 & 586 & 6.7 \\
        AMIR~\cite{Sadeghian2017TrackingTU} & Online & 37.6 &\color{red}46.0  & 71.7 & 15.8\% & \textbf{26.8\%} & 7933	 & 29397	& 1026 & 1.9 \\
         RAR15pub~\cite{Kuan2017Recurrent}& Online & 35.1 &45.4  & 70.9 & 13.0\% & 42.3\% & 6771	& 32717	   & 381 & 5.4 \\
        DCCRF~\cite{DCCRFZhou2018Deep} & Online & 33.6 &39.1 & 70.9 & 10.4 \% & 37.6\% & 5917	 & 34002	& 866  &0.1\\
%TDAM~\cite{TDAMYang2016TemporalDA} & Online & 33.00 &{\color{red}72.80} & 13.30\% & 39.10\% & 10064 & 30617 & 464 & 5.9\\
    CDA~\cite{ConfidenceBae2014RobustOM} & Online & 32.8 &38.8 & 70.7 & 9.7\% & 42.2\% & \color{red}4983 & 35690 & 614 & 2.3 \\
    MDP~\cite{DecisionMakingXiang2015LearningTT} & Online &30.3 & 44.7 & 71.3 & 13.0\% &38.4\% & 9717 &32422 &680 &1.1\\
%SCEA~\cite{SCEAYoon2016OnlineMT} & Online & 29.10  &71.10 & 8.90\% & 47.30\% & 6060 &36912 &604 & 6.8\\
%oICF~\cite{oICFKieritz2016OnlineMT} & Online & 27.10 & 70.00 & 6.40\% & 48.70\% & 7594 & 36757 & 4545 & 1.4 \\
    RNNLSTM~\cite{OnlineRNNMilan2017OnlineMT} & Online &19.0 &17.1 &71.0 & 5.5\% &45.6\% & 11578 & 36706 & 1490 & \textbf{165}\\
%RMOT~\cite{RMOTYoon2015BayesianMT} & Online & 18.60 & 69.60 & 5.30\% & 53.30\% & 12473 & 36835 & 684 & 7.9 \\
%TC-ODAL~\cite{TCODALBae2014RobustOM} & Online & 15.10 & 70.50 & 3.20\% & 55.80\% & 12970 & 38538 & 637 & 1.7 \\
\hline
\end{tabular}
\end{center}
\end{table*}

%	We conduct experiments on the 2DMOT2017 dataset. Table \ref{tab:table10} shows that our method achieves the best performance in terms of MOTA (53.1\%), MT(24.9\%),ML(30.7\%) and FN (234991).
%	
%	\begin{table}
%		\caption{Result on the MOT2017 test dataset. The best and second best results are highlighted by bold and red fonts..}
%		\label{tab:table10}
%		\vspace{-2mm}
%		\begin{center}
%			\resizebox{82mm}{10mm}{
%				\begin{tabular}{|l|ccccccc|}
%					\hline
%					Tracker & MOTA$\uparrow$ & MOTP$\uparrow$ & MT$\uparrow$ & ML$\downarrow$ & FP$\downarrow$ & FN$\downarrow$ & IDS $\downarrow$ \\
%					\hline
%					\textbf{Ours}  & \textbf{53.1} & 76.1 & \textbf{24.2\%} & \textbf{30.7\%} & 27194 & \textbf{234991} & 2518  \\
%					eHAF17~\cite{Sheng2018Heterogeneous} & \color{red}51.8 & \textbf{77.0} & \color{red}23.4\% & 37.9\% & 33212 & \color{red}236772 & \color{red}1834 \\
%					FWT~\cite{Henschel2018Fusion}  & 51.3 & 77.0 & 21.4\% & \color{red}35.2\% & \color{red}24101 & 247921 & 2648  \\
%					JCC~\cite{KM2018Multi}         & 51.2 &75.9 & 20.9\% &37.0\%    &25937 &247822   &\textbf{1802} \\
%					MHT-DAM~\cite{NHTKim2015MultipleHT}  & 50.7 &\textbf{77.5} & 20.8\% &36.9\% &\textbf{22875} &252889 &2314 \\
%					\hline
%			\end{tabular}}
%		\end{center}
%	\vspace{-2mm}
%	\end{table}
%	
	
	\section{Conclusion}
In this paper, we present a deep learning based CRF framework for multi-object tracking. To exploit the dependencies between detection results, we pay more attention to handle difficult node pairs when modeling pairwise potentials. Specifically, we use bidirectional LSTMs to solve this joint probability matching problem. We pose the CRF inference  as an RNN learning process using the standard gradient descent algorithm, where unary and pairwise potentials are jointly optimized in an end-to-end manner.
%Extensive experiments on three standard benchmark datasets demonstrate the effectiveness of the proposed method.
Extensive experimental results on the challenging MOT datasets, including MOT-2015 and MOT-2016 demonstrate the effectiveness of the proposed method that achieves new state-of-the-art performance on both datasets.
 %Note that a potential (promising) solution to the optimization problems is Graph Neural Networks, which are able to learn locally adaptive pairwise edge potentials and would therefore offer a mathematical foundation alternative (compared) to the proposed RNN based optimization.

{\small
\bibliographystyle{ieee}
\bibliography{egbib}

\begin{thebibliography}{10}\itemsep=-1pt

\bibitem{ConfidenceBae2014RobustOM}
S.~H. Bae and K.-J. Yoon.
\newblock Robust online multi-object tracking based on tracklet confidence and
  online discriminative appearance learning.
\newblock {\em 2014 IEEE Conference on Computer Vision and Pattern
  Recognition}, pages 1218--1225, 2014.

\bibitem{Ban2016Tracking}
Y.~Ban, S.~Ba, X.~Alameda-Pineda, and R.~Horaud.
\newblock {\em Tracking Multiple Persons Based on a Variational Bayesian
  Model}.
\newblock 2016.

\bibitem{KShortestBerclaz2011MultipleOT}
J.~Berclaz, F.~Fleuret, E.~T{\"u}retken, and P.~Fua.
\newblock Multiple object tracking using k-shortest paths optimization.
\newblock {\em IEEE Transactions on Pattern Analysis and Machine Intelligence},
  33:1806--1819, 2011.

\bibitem{PrincipledIntegrationBeyer2017TowardsAP}
L.~Beyer, S.~Breuers, V.~Kurin, and B.~Leibe.
\newblock Towards a principled integration of multi-camera re-identification
  and tracking through optimal bayes filters.
\newblock {\em 2017 IEEE Conference on Computer Vision and Pattern Recognition
  Workshops (CVPRW)}, pages 1444--1453, 2017.

\bibitem{Bo2014Multi}
Y.~Bo and R.~Nevatia.
\newblock Multi-target tracking by online learning a crf model of appearance
  and motion patterns.
\newblock {\em International Journal of Computer Vision}, 107(2):203--217,
  2014.

\bibitem{Near-OnlineChoi2015NearOnlineMT}
W.~Choi.
\newblock Near-online multi-target tracking with aggregated local flow
  descriptor.
\newblock {\em 2015 IEEE International Conference on Computer Vision (ICCV)},
  pages 3029--3037, 2015.

\bibitem{Chu2019Online}
P.~Chu, H.~Fan, C.~C. Tan, and H.~Ling.
\newblock Online multi-object tracking with instance-aware tracker and dynamic
  model refreshment.
\newblock In {\em 2019 IEEE Winter Conference on Applications of Computer
  Vision (WACV)}, 2019.

\bibitem{Kuan2017Recurrent}
K.~Fang, Y.~Xiang, X.~Li, and S.~Savarese.
\newblock Recurrent autoregressive networks for online multi-object tracking.
\newblock pages 466--475, 2017.

\bibitem{Heili2011Detection}
A.~Heili, C.~Chen, and J.~M. Odobez.
\newblock Detection-based multi-human tracking using a crf model.
\newblock In {\em IEEE International Conference on Computer Vision Workshops},
  pages 1673--1680, 2011.

\bibitem{Heili2014Exploiting}
A.~Heili, A.~L¨®pezm¨¦ndez, and J.~M. Odobez.
\newblock Exploiting long-term connectivity and visual motion in crf-based
  multi-person tracking.
\newblock {\em IEEE Transactions on Image Processing A Publication of the IEEE
  Signal Processing Society}, 23(7):3040--3056, 2014.

\bibitem{hermans2017defense}
A.~Hermans, L.~Beyer, and B.~Leibe.
\newblock In defense of the triplet loss for person re-identification.
\newblock {\em arXiv preprint arXiv:1703.07737}, 2017.

\bibitem{DefenseTripHermans2017InDO}
A.~Hermans, L.~Beyer, and B.~Leibe.
\newblock In defense of the triplet loss for person re-identification.
\newblock {\em CoRR}, abs/1703.07737, 2017.

\bibitem{Hochreiter1997Long}
S.~Hochreiter and J.~Schmidhuber.
\newblock Long short-term memory.
\newblock {\em Neural Computation}, 9(8):1735--1780, 1997.

\bibitem{HierarchicalHuang2008RobustOT}
C.~Huang, B.~Wu, and R.~Nevatia.
\newblock Robust object tracking by hierarchical association of detection
  responses.
\newblock In {\em ECCV}, 2008.

\bibitem{Keuper2018Motion}
M.~Keuper, S.~Tang, B.~Andres, T.~Brox, and B.~Schiele.
\newblock Motion segmentation and multiple object tracking by correlation
  co-clustering.
\newblock {\em IEEE Transactions on Pattern Analysis and Machine Intelligence},
  PP(99):1--1, 2018.

\bibitem{NHTKim2015MultipleHT}
C.~Kim, F.~Li, A.~Ciptadi, and J.~M. Rehg.
\newblock Multiple hypothesis tracking revisited.
\newblock {\em 2015 IEEE International Conference on Computer Vision (ICCV)},
  pages 4696--4704, 2015.

\bibitem{AdamOpKingma2014AdamAM}
D.~P. Kingma and J.~Ba.
\newblock Adam: A method for stochastic optimization.
\newblock {\em CoRR}, abs/1412.6980, 2014.

\bibitem{discriminativeAppKuo2010MultitargetTB}
C.-H. Kuo, C.~Huang, and R.~Nevatia.
\newblock Multi-target tracking by on-line learned discriminative appearance
  models.
\newblock {\em 2010 IEEE Computer Society Conference on Computer Vision and
  Pattern Recognition}, pages 685--692, 2010.

\bibitem{ArbitraryLarsson2017LearningAP}
M.~Larsson, F.~Kahl, S.~Zheng, A.~Arnab, P.~H.~S. Torr, and R.~I. Hartley.
\newblock Learning arbitrary potentials in crfs with gradient descent.
\newblock {\em CoRR}, abs/1701.06805, 2017.

\bibitem{CRF-BasedLe2016LongTermTC}
N.~Le, A.~Heili, and J.-M. Odobez.
\newblock Long-term time-sensitive costs for crf-based tracking by detection.
\newblock In {\em ECCV Workshops}, 2016.

\bibitem{siamesecnn2016LearningBT}
L.~Leal-Taix{\'e}, C.~Canton-Ferrer, and K.~Schindler.
\newblock Learning by tracking: Siamese cnn for robust target association.
\newblock {\em 2016 IEEE Conference on Computer Vision and Pattern Recognition
  Workshops (CVPRW)}, pages 418--425, 2016.

\bibitem{leal2015motchallenge}
L.~Leal-Taix{\'e}, A.~Milan, I.~Reid, S.~Roth, and K.~Schindler.
\newblock Motchallenge 2015: Towards a benchmark for multi-target tracking.
\newblock {\em arXiv preprint arXiv:1504.01942}, 2015.

\bibitem{networkflowsLin2008GlobalDA}
X.~Lin, Y.~Li, and R.~Nevatia.
\newblock Global data association for multi-object tracking using network
  flows.
\newblock {\em 2008 IEEE Conference on Computer Vision and Pattern
  Recognition}, pages 1--8, 2008.

\bibitem{Long2018Online}
C.~Long, H.~Ai, S.~Chong, Z.~Zhuang, and B.~Bo.
\newblock Online multi-object tracking with convolutional neural networks.
\newblock In {\em 2017 IEEE International Conference on Image Processing
  (ICIP)}, 2018.

\bibitem{Ma2018Trajectory}
C.~Ma, C.~Yang, F.~Yang, Y.~Zhuang, Z.~Zhang, H.~Jia, and X.~Xie.
\newblock Trajectory factory: Tracklet cleaving and re-connection by deep
  siamese bi-gru for multiple object tracking.
\newblock 2018.

\bibitem{MaTangACCV2018}
L.~Ma, S.~Tang, M.~J. Black, and L.~V. Gool.
\newblock Customized multi-person tracker.
\newblock In {\em Computer Vision -- ACCV 2018}. Springer International
  Publishing, Dec. 2018.

\bibitem{milan2016mot16}
A.~Milan, L.~Leal-Taix{\'e}, I.~Reid, S.~Roth, and K.~Schindler.
\newblock Mot16: A benchmark for multi-object tracking.
\newblock {\em arXiv preprint arXiv:1603.00831}, 2016.

\bibitem{OnlineRNNMilan2017OnlineMT}
A.~Milan, S.~H. Rezatofighi, A.~R. Dick, K.~Schindler, and I.~D. Reid.
\newblock Online multi-target tracking using recurrent neural networks.
\newblock In {\em AAAI}, 2017.

\bibitem{Discrete-Continuous2016MultiTargetTB}
A.~Milan, K.~Schindler, and S.~Roth.
\newblock Multi-target tracking by discrete-continuous energy minimization.
\newblock {\em IEEE Transactions on Pattern Analysis and Machine Intelligence},
  38:2054--2068, 2016.

\bibitem{Globally-optimalgreedy2011GloballyoptimalGA}
H.~Pirsiavash, D.~Ramanan, and C.~C. Fowlkes.
\newblock Globally-optimal greedy algorithms for tracking a variable number of
  objects.
\newblock {\em CVPR 2011}, pages 1201--1208, 2011.

\bibitem{Sadeghian2017TrackingTU}
A.~Sadeghian, A.~Alahi, and S.~Savarese.
\newblock Tracking the untrackable: Learning to track multiple cues with
  long-term dependencies.
\newblock {\em 2017 IEEE International Conference on Computer Vision (ICCV)},
  pages 300--311, 2017.

\bibitem{Sanchez2016Online}
R.~Sanchez-Matilla, F.~Poiesi, and A.~Cavallaro.
\newblock {\em Online Multi-target Tracking with Strong and Weak Detections}.
\newblock 2016.

\bibitem{ShengIterative}
H.~Sheng, J.~Chen, Y.~Zhang, W.~Ke, Z.~Xiong, and J.~Yu.
\newblock Iterative multiple hypothesis tracking with tracklet-level
  association.
\newblock {\em IEEE Transactions on Circuits and Systems for Video Technology}.

\bibitem{Hao2018Heterogeneous}
H.~Sheng, Y.~Zhang, J.~Chen, Z.~Xiong, and J.~Zhang.
\newblock Heterogeneous association graph fusion for target association in
  multiple object tracking.
\newblock {\em IEEE Transactions on Circuits and Systems for Video Technology},
  PP(99):1--1, 2018.

\bibitem{Large-Scale2014VeryDC}
K.~Simonyan and A.~Zisserman.
\newblock Very deep convolutional networks for large-scale image recognition.
\newblock {\em CoRR}, abs/1409.1556, 2014.

\bibitem{QuadrupletSon2017MultiobjectTW}
J.~Son, M.~Baek, M.~Cho, and B.~Han.
\newblock Multi-object tracking with quadruplet convolutional neural networks.
\newblock {\em 2017 IEEE Conference on Computer Vision and Pattern Recognition
  (CVPR)}, pages 3786--3795, 2017.

\bibitem{LiftedMulticutTang2017MultiplePT}
S.~Tang, M.~Andriluka, B.~Andres, and B.~Schiele.
\newblock Multiple people tracking by lifted multicut and person
  re-identification.
\newblock {\em 2017 IEEE Conference on Computer Vision and Pattern Recognition
  (CVPR)}, pages 3701--3710, 2017.

\bibitem{ConstrainedWang2016JointLO}
B.~Wang, K.~L. Chan, L.~Wang, B.~Shuai, Z.~Zuo, T.~Liu, and G.~Wang.
\newblock Joint learning of convolutional neural networks and temporally
  constrained metrics for tracklet association.
\newblock {\em 2016 IEEE Conference on Computer Vision and Pattern Recognition
  Workshops (CVPRW)}, pages 386--393, 2016.

\bibitem{CoherentDynamicsWang2017TrackletAB}
B.~Wang, G.~Wang, K.~L. Chan, and L.~Wang.
\newblock Tracklet association by online target-specific metric learning and
  coherent dynamics estimation.
\newblock {\em IEEE Transactions on Pattern Analysis and Machine Intelligence},
  39:589--602, 2017.

\bibitem{HFRFXiang2016MultitargetTU}
J.~Xiang, N.~Sang, J.~Hou, R.~Huang, and C.~Gao.
\newblock Multitarget tracking using hough forest random field.
\newblock {\em IEEE Transactions on Circuits and Systems for Video Technology},
  26:2028--2042, 2016.

\bibitem{bmvcXiang2018MultipleOT}
J.~Xiang, G.~Zhang, N.~Sang, R.~Huang, and J.~Hou.
\newblock Multiple object tracking by learning feature representation and
  distance metric jointly.
\newblock In {\em BMVC}, 2018.

\bibitem{DecisionMakingXiang2015LearningTT}
Y.~Xiang, A.~Alahi, and S.~Savarese.
\newblock Learning to track: Online multi-object tracking by decision making.
\newblock {\em 2015 IEEE International Conference on Computer Vision (ICCV)},
  pages 4705--4713, 2015.

\bibitem{Learningaffinities2011LearningAA}
B.~Yang, C.~Huang, and R.~Nevatia.
\newblock Learning affinities and dependencies for multi-target tracking using
  a crf model.
\newblock {\em CVPR 2011}, pages 1233--1240, 2011.

\bibitem{onlineCRFYang2012AnOL}
B.~Yang and R.~Nevatia.
\newblock An online learned crf model for multi-target tracking.
\newblock {\em 2012 IEEE Conference on Computer Vision and Pattern
  Recognition}, pages 2034--2041, 2012.

\bibitem{CUHK03Zheng2015ScalablePR}
L.~Zheng, L.~Shen, L.~Tian, S.~Wang, J.~Wang, and Q.~Tian.
\newblock Scalable person re-identification: A benchmark.
\newblock {\em 2015 IEEE International Conference on Computer Vision (ICCV)},
  pages 1116--1124, 2015.

\bibitem{CRFasRNNZheng2015ConditionalRF}
S.~Zheng, S.~Jayasumana, B.~Romera-Paredes, V.~Vineet, Z.~Su, D.~Du, C.~Huang,
  and P.~H.~S. Torr.
\newblock Conditional random fields as recurrent neural networks.
\newblock {\em 2015 IEEE International Conference on Computer Vision (ICCV)},
  pages 1529--1537, 2015.

\bibitem{DCCRFZhou2018Deep}
H.~Zhou, W.~Ouyang, J.~Cheng, X.~Wang, and H.~Li.
\newblock Deep continuous conditional random fields with asymmetric
  inter-object constraints for online multi-object tracking.
\newblock {\em IEEE Transactions on Circuits and Systems for Video Technology},
  PP(99):1--1, 2018.

\bibitem{Zhu2019Online}
J.~Zhu, H.~Yang, N.~Liu, M.~Kim, W.~Zhang, and M.-H. Yang.
\newblock Online multi-object tracking with dual matching attention networks.
\newblock 2019.

\end{thebibliography}
}

\end{document}